\documentclass{article}

\usepackage{microtype}
\usepackage{amsmath}
\usepackage{graphicx}
\usepackage{subfig}

\usepackage{booktabs} 

\usepackage{hyperref}

\usepackage[accepted]{icml2019}

\icmltitlerunning{BERT and PALs: Projected Attention Layers for
           Efficient Adaptation in Multi-Task Learning}

\begin{document}

\twocolumn[
\icmltitle{BERT and PALs: Projected Attention Layers for \\
           Efficient Adaptation in Multi-Task Learning}

\begin{icmlauthorlist}

\icmlauthor{Asa Cooper Stickland}{ed}
\icmlauthor{Iain Murray}{ed}

\end{icmlauthorlist}

\icmlaffiliation{ed}{School of Informatics, University of Edinburgh}

\icmlcorrespondingauthor{Asa Cooper Stickland}{a.cooper.stickland@ed.ac.uk}

\icmlkeywords{Machine Learning, ICML}

\vskip 0.3in
]

\printAffiliationsAndNotice{}  

\begin{abstract}
Multi-task learning shares information
between related tasks, sometimes reducing the number of parameters required. State-of-the-art results across multiple natural language understanding tasks in the GLUE benchmark have previously used transfer from a single large task: unsupervised pre-training with BERT, where a separate BERT model was fine-tuned for each task. We explore multi-task approaches that share a \hbox{single} BERT model with a small number of additional task-specific parameters. Using new adaptation modules, PALs or `projected attention layers', we match the performance of separately fine-tuned models on the GLUE benchmark with $\approx$7 times fewer parameters, and obtain state-of-the-art results on the Recognizing Textual Entailment dataset.
\end{abstract}

\section{Introduction}
\label{submission}

This work explores how to adapt a single large base model to work with multiple tasks. In particular we focus on using deep neural networks, pre-trained on large amounts of English text, for multi-task learning on several natural language understanding (NLU) tasks.

Some multi-task learning approaches consider learning a general-purpose model that shares all parameters across tasks \citep[e.g., the NLP decathlon introduced by][]{deca}. This setting requires all tasks to have the same input and output space, and the input indicates the task. Instead, we consider the setting where we share most parameters across all tasks, but have a small number of task-specific parameters which adapt the shared model.

Sharing parameters, and thus a common representation, between tasks can sometimes lead to better generalization. However, fine-tuning separate models for each task often works better in practice. Although we are interested in multi-task methods that give results close to (or better than) state-of-the-art, there are separate motivations for maintaining shared parameters between tasks:
\begin{itemize}

 \item On applications like mobile devices we may have constraints on battery life. Applying several different neural networks to the same input costs energy. If only the `tops' of our models are task-specific, we can apply a shared transformation only once to the input, and use this transformed representation multiple times, as input to each task-specific function.
 \item  Again on mobile devices, running several different neural networks for various tasks
 can incur a computational and energy overhead due to swapping parameters on a dedicated integrated circuit \citep{adapt}.
    \item An application with a large number of tasks may have constraints on the number of parameters that can be stored. For example, web-scale applications may need to avoid storing a separate large model for every user. 
\end{itemize}

Given a large number of shared parameters in a base model, and a small number of task-specific parameters, our key questions are: where should we be transforming the base model? What form should these transformations take? We assume the task is always known, so the model can always choose the correct adaptation parameters and output space.

We experiment on a set of eight NLU tasks from the GLUE benchmark \citep{glue},
which include question answering, sentiment analysis, and textual entailment.
The number of training examples varies widely across the tasks, so we explore how
to schedule training to not unduly favor the well-resourced tasks, or overfit
the low-resource tasks.

We use the BERT model \citep[Bidirectional Encoder Representations from Transformers,][]{bert} as our base pre-trained model. Pre-trained BERT representations can be fine-tuned with just one additional output layer to create state-of-the-art models for a wide range of tasks, including the GLUE benchmark. However, the entire model is fine-tuned, meaning we need a separate model for each task. The transformer architecture that BERT is based on is powerful and popular, so finding the best way to adapt the parameters of this architecture for multi-task learning may be useful in other contexts, such as multilingual machine translation.

Our main contributions are:
1)~We introduce the `Projected Attention Layer' (PAL), a low-dimensional multi-head attention layer that is added in parallel to normal BERT layers.
2)~We introduce a novel method for scheduling training, where we sample tasks proportional to their training set size at first, and de-emphasize training set size as training proceeds.
3)~We perform an empirical comparison of alternative adaptation modules for self-attention-based architectures.

Making links to the vision literature, we identify shared lessons for where to add task-adaptation parameters depending on resource constraints.
On the GLUE benchmark, we show that PALs enable comparable performance to fine-tuned BERT-base (the smaller of the two models considered by \citealt{bert}) on many tasks with $\approx$7 times fewer parameters. We improve the performance of BERT-base on the recognising textual entailment (RTE) task, achieving 76.6\% accuracy, surpassing the performance of fine-tuned BERT-large (70.1\%) and the MT-DNN model \cite{mtdnn} (75.5\%) which also uses BERT and multi-task learning. We also find that the more parameter sharing we have, the better we do on the RTE task.

\section{Background}
Multi-task learning aims to provide an inductive bias that means models have
to learn features that are general enough to perform well on many tasks \citep{Caruana}. In NLP, examples of previous work include using a single model for chunking, tagging, named entity recognition, and semantic role labeling by applying a shared neural network to text, with different output layers \citep{Collobert}. Another approach outputs predictions at different layers using the idea of a linguistic hierarchy \citep{hash, hmtl}. \citet{sub} train a sequence-to-sequence RNN model on tasks including machine translation and natural language inference, and learn sentence representations useful for downstream tasks. Outside NLP, multi-task learning has been applied to diverse domains such as speech recognition \citep{deng} and reinforcement learning \citep{rein}. \citet{ruder} provides a more general overview.

Many multi-task learning approaches can be categorized as either `hard parameter sharing' or `soft parameter sharing'. Hard parameter sharing uses the same hidden layers for all tasks, with task-specific output layers.
Soft parameter sharing gives each task its own model, but the distances between the parameters of the models are regularized to encourage the parameters to be similar. For example \citet{duong} use the L2 distance, and \citet{yang} use the trace norm. In this work we assume that soft-parameter sharing with the whole of BERT requires too many parameters. We instead explore how to do hard-parameter sharing, by adding adapters to shared layers, as well as the usual separate output layers.

\subsection{Adaptation Parameters}
\label{adapt}

Various strategies for adding adaptation parameters have been explored. \emph{Learning hidden unit contributions} \citep[LHUC,][]{7078569} modifies a neural network by multiplying each hidden unit by a learnable scalar. Since the number of units is much smaller than the number of parameters in the network, this approach adds a small number of parameters compared to other methods we consider.

\emph{Residual adapter modules} \citep{adapt} adapt large pre-trained residual networks \citep{resnet} for multi-task learning in computer vision. 
Each adapter module contains a 1$\times$1 filter bank with a skip connection, which can be inserted \emph{in series}, between the original network layers, or \emph{in parallel}, as additional inputs to a layer. For a layer with $C$ channels, the module contains an additional $C \!\times\! C$ matrix per layer for each task, containing $C$ 1$\!\times\!$1 convolutional filters.
This $C \!\times\! C$ matrix can be compressed by replacing it with a low-rank approximation, so that the
adapters contain a small fraction of the model parameters (e.g., less than 10\% for each task).
Several of our methods were inspired by the idea of using a low-rank approximation to the key operation of a model: the convolutional layer when dealing with images, or multi-head attention in the transformer.

\subsection{Fine-tuning Approaches}

A recent trend in transfer learning is to pre-train some model architecture
on a language
modeling objective before fine-tuning
that same model for a supervised downstream
task \citep{dai,ulmft,gpt}. BERT uses a similar approach, but was pre-trained with two objectives: 1)~filling in words `masked' out of an input sentence, and 2)~classifying whether two input sentences are adjacent in a corpus of text. Unlike a normal language modeling objective, BERT conditions on both left and right context when predicting the masked words.

The neural network layers in BERT are taken from the Transformer model \citep{NIPS2017_7181}, a sequence to sequence model that achieved state-of-the-art results in machine translation. Transformer layers have subsequently been used more broadly, e.g.\ for language modeling \citep{xl}, image generation \citep{sagan}, and generalized to video classification, object detection/segmentation and human pose estimation \citep{NonLocal2018}.

A concurrent approach by \citet{bertadapt}, introduces adapters similar to our `low-rank' layers (section~\ref{sec:within}), but added within each layer before each application of layer-norm. This work also keeps the BERT model fixed while training adapter modules. We concentrated on jointly fine-tuning the entire BERT model on all tasks, which has downsides: 1)~interference and `forgetting' of stored knowledge is possible; 2)~we require access to all tasks at training time. However the multi-task setup requires less adaptation parameters for good performance (we use 1.13$\times$ parameters compared to their 1.3$\times$ parameters\footnote{Although the results are not directly comparable since \citet{bertadapt} use BERT-large and we use BERT-base.} to match having separate models for each GLUE task.), and is crucial for the transfer effects that gave us good performance on RTE.

\section{Adapting Self Attention}
The BERT model we are adapting is a multi-layer bidirectional Transformer encoder based on the original model of \citet{NIPS2017_7181}. We only consider the smaller BERT-base model, which contains 110 million parameters.
We somewhat arbitrarily limit ourselves to a 1.13$\times$ increase in total parameters, which is equivalent to 15 million, or 1.9 million parameters per task. This choice avoids the extremes of having nearly no extra task-specific parameters, or giving each task its own whole model.

In the following sections we first introduce various components of the full BERT model, and discuss how many parameters they require (section~\ref{sec:param}). We then show the exact form our parameter additions took, distinguishing between adding to the `top' of the model, just before the output space (section~\ref{sec:top}),  or within each layer of the BERT-base architecture (section~\ref{sec:within}).
\subsection{Model Architecture and Multi-head Attention}
\label{sec:param}

BERT takes in a sequence (one or two English sentences in our case) and outputs a vector representation of that sequence. Each token in the sequence has its own hidden vector, and the first token of every sequence is always a special classification embedding (\texttt{[CLS]}). At each layer of BERT the hidden states of every sequence element are transformed, but only the final hidden state of \texttt{[CLS]} is used for classification/regression tasks. We now describe how the vector for one element of the sequence is transformed.

The multi-head attention layer \citep{NIPS2017_7181} is the core of the transformer architecture that transforms hidden states for each element of a sequence based on the other elements (the fully-connected layers act on each element separately). The multi-head layer, which we write as $\mathrm{MH}(\cdot)$, consists of $n$ different dot-product attention mechanisms. At a high level, attention represents a sequence element with a weighted sum of the hidden states of all the sequence elements. In multi-head attention the weights in the sum use dot product similarity between transformed hidden states.

Concretely, the $i$th attention mechanism `head' is:
\begin{equation}
    \mathrm{Attention}_i(\mathbf{h}_j) = \sum_t \mathrm{softmax}\bigg(\frac{W^q_i \mathbf{h}_j\cdot W^k_i \mathbf{h}_t}{\sqrt{d/n}}\bigg) W^v_i \mathbf{h}_t 
\end{equation}
where $\mathbf{h}_j$ (we drop the $j$ index in the following discussion) is a $d$ dimensional hidden vector for a particular sequence element, and $t$ runs over every sequence element. In BERT the $W^q_i$, $W^k_i$ and $W^v_i$ are matrices of size $d/n \times d$, and so each `head' projects down to a different subspace of size $d/n$, attending to different information. 
Finally the outputs of the $n$ attention heads (each of size $d/n$) are concatenated together (which we show as $[\cdot,...,\cdot]$) and linearly transformed:
\begin{equation}
    \mathrm{MH}(\mathbf{h}) = W^o \,[\mathrm{Attention}_1(\mathbf{h}),...,\mathrm{Attention}_n(\mathbf{h})]
\end{equation}
with $W^o$ a $d \times d$ matrix\footnote{\citet{NIPS2017_7181} provide a more detailed motivation and discussion.}. 
Throughout this section, we ignore terms linear in $d$ (like bias terms) to avoid clutter, as they don't add significantly to the parameter count. The matrices in a multi-head layer have $3 n d^2/n + d^2 = 4d^2$ parameters.

We further define another component of a BERT layer, the self-attention layer, which we write as $\mathrm{SA}(\cdot)$:
\begin{equation}
    \mathrm{SA}(\mathbf{h}) = \mathrm{FFN}(\mathrm{LN}(\mathbf{h} + \mathrm{MH}(\mathbf{h}))),
\end{equation}
LN($\cdot$) is \emph{layer normalisation} \citep{Ba2016LayerN}, requiring $2d$ parameters.
FFN is a standard \emph{feed-forward network},
\begin{equation}
    \mathrm{FFN}(\mathbf{h}) = W_2f(W_1\mathbf{h} + b_1) + b_2,
\end{equation}
with $f(\cdot)$ a non-linearity, GeLU \citep{gelu} in BERT\@. Matrix $W_1$ has size $d_{ff} \times d$ and $W_2$ has size $d \times d_{ff}$, so overall we require $2d d_{ff}$ parameters from the FFN component. 

Putting this together, a BERT layer, which we write BL$(\cdot)$, is layer-norm applied to the output of a self-attention layer, with a residual connection.
\begin{equation}
    \mathrm{BL}(\mathbf{h}) = \mathrm{LN}(\mathbf{h} + \mathrm{SA}(\mathbf{h}) )
\end{equation}
We have $4d^2 + 2dd_{ff}$ total parameters from a BERT layer.

The entire BERT model is simply a stack of 12 BERT layers, followed by (in our case) a transformation to take us to the output space for a NLU task. We write the dimensions of the hidden states in BERT-base as $d_m\!=\!768$. The final hidden state of the first token of every sequence is all that is used for the transformation to the output.

The exact form of the transformation applied to the final hidden state of the \texttt{[CLS]} token is a simple $d \times d$ linear transformation, known as a \textbf{`pooling layer'}, followed by a nonlinearity then another matrix multiply that projects to the output space. The output space is always three dimensional or less in our case, and so this projection does not require many parameters. However separate pooling layers add $d^2$ parameters for each task. When sharing this layer we needed to use a non-standard training schedule; see section~\ref{samp}.

\subsection{Adding Parameters to the Top}
\label{sec:top}
The simplest way to add parameters to a model is to add them at the `top' of the model, i.e.\ just before the classification layer.

We get our final hidden state for \texttt{[CLS]}, $\mathbf{h}^{f}$, from the original vector embeddings of the tokens in the sequence (of length $l$), $\{\mathbf{h}_t\}_{t=0}^{l}$, by
\begin{equation}
    \mathbf{h}^{f} = \mathrm{TS}( \mathrm{BERT}(\{\mathbf{h}_t\}_{t=0}^{l} )),
\end{equation}
where $\mathrm{TS}(\cdot)$ is a task-specific function that can potentially operate on a single vector, but depends on the entire sequence when it contains attention layers. $\mathrm{BERT}(\cdot)$ always depends on the entire sequence, and is shared across tasks.

The benefits of this form are that at inference time we only apply $\mathrm{BERT}(\{\mathbf{h}_t\}_{t=0}^{l})$ once (assuming the setting where we perform multiple tasks on the same piece of text), which saves significantly on total operations because each $\mathrm{TS}(\cdot)$ requires much fewer operations than the main BERT model.

The simplest form for the task-specific transformation of the hidden state $\mathrm{TS}(\cdot)$ would be a linear transform followed by a nonlinearity. However this requires $d_m^2$ parameters, and $d_m$ is fairly large even for BERT-base. The linear transform does not violate our 15 million parameter constraint, but we expect there are more efficient ways to add parameters.

Another obvious transformation, adding an extra BERT layer for each task, results in approximately a 1.67$\times$ increase in number of parameters, or 73 million new parameters. $d_{ff}$ is $4d_m$ for BERT, so for a BERT layer we get $4d_m^2 + 2d_m d_{ff} = 12d_m^2$ parameters. We include this architecture in our experiments for comparison, with the caveat that it requires many more parameters than our alternatives.

To avoid transformations requiring $O(d_m^2)$ parameters, we propose using task-specific functions of the form
\begin{equation}
    \mathrm{TS}(\mathbf{h}) = V^D g(V^E \mathbf{h}),
\end{equation}
where $V^E$ is a $d_s \!\times\! d_m$ `encoder' matrix, $V^D$ is a $d_m \!\times\! d_s$ `decoder' matrix with $d_s \!<\! d_m$, and $g(\cdot)$ is an arbitrary function. Because we can make $d_s$ as small as we like, $g(\cdot)$ can be composed of multiple layers of transformations, and not impose a large parameter budget.

We experiment with these choices for each layer of $g(\cdot)$:
\begin{itemize}
    \item Multi-head attention, optionally followed by a residual connection and layer-norm. We refer to this method as \textbf{Projected Attention}. We found $d_s=204$ worked well,
and allowed us to stay within our $1.13\times$ parameter limit.
    \item A one or two layer feed-forward network followed by a residual connection and layer-norm, such that it has the same number of parameters as the previous form; this means the intermediate layer is of size 408 (for a one layer network) or 252 (for a two layer network).
\end{itemize}

\begin{figure}[t]
\begin{center}
\centerline{\includegraphics[width=1.0in]{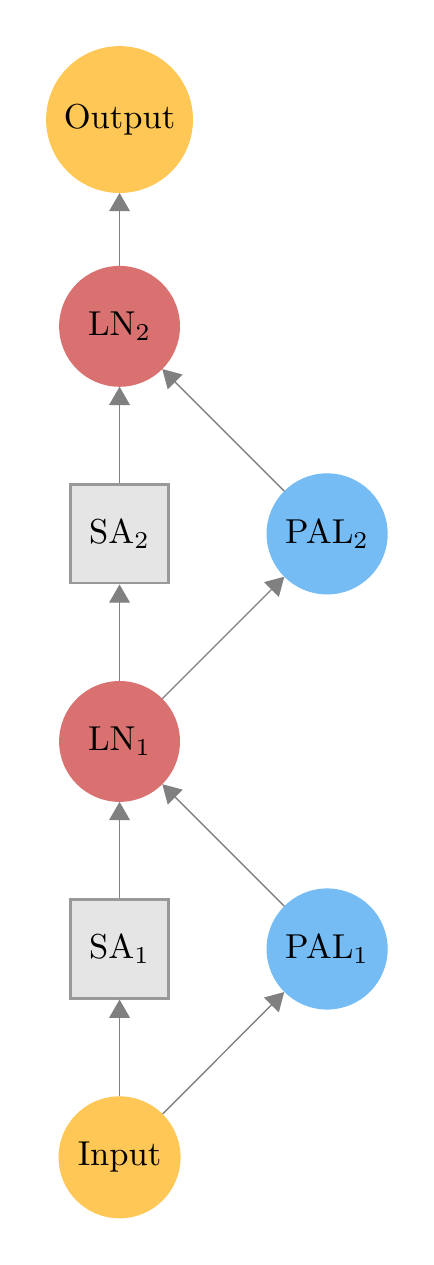}}
\vskip -0.1in
\caption{Schematic diagram of adding a task-specific function (here our `Projected Attention Layers' or PALs) in parallel with self-attention (SA) layers in a BERT model (see section~\ref{sec:within}), with only two layers for simplicity. LN refers to layer-norm.}
\label{bert}
\end{center}
\vskip -0.2in
\end{figure}
\subsection{Adding Parameters within BERT}
\label{sec:within}
Instead of adding parameters to the top of the model, we may want to modify the $\mathrm{BERT}(\cdot)$ function itself, inspired by `residual adapter modules' (section~\ref{adapt}, \citealp{adapt}). Specifically, we wish to add task-specific parameters to each layer of the BERT model. See figure~\ref{bert} for an illustration.

We can add a task-specific function `in parallel' with each BERT layer as follows:
\begin{equation}
    \mathbf{h}^{l+1} = \mathrm{LN}(\mathbf{h}^l + \mathrm{SA}(\mathbf{h}^l) + \mathrm{TS}(\mathbf{h}^l))
\end{equation}
where $l$ indexes the layer. This means we recover the original BERT model if $\mathrm{TS}(\cdot)$ outputs a zero vector.
Alternatively we can add a `serial' connection where we transform the output of a BERT layer:
\begin{equation}
 \mathbf{\hat{h}}^{l+1} = \mathrm{LN}(\mathbf{h}^l + \mathrm{SA}(\mathbf{h}^l))
\end{equation}
\begin{equation}
\mathbf{h}^{l+1} = \mathrm{LN}(\mathbf{\hat{h}}^{l+1} + \mathrm{TS}(\mathbf{\hat{h}}^{l+1})).
\end{equation}
  In preliminary experiments, serial connections gave consistently much worse results than parallel connections, and we report results for parallel connections in what follows.

  We again consider task-specific functions of the form: \begin{equation}
    \mathrm{TS}(\mathbf{h}) = V^D g(V^E \mathbf{h}),
\end{equation}
with the difference that $V^E$ (again a $d_s \times d_m$ matrix with $d_s < d_m$) and $V^D$ (again a $d_m \times d_s$ matrix) are needed at each layer rather than only once each.

We experiment with $g(\cdot)$ taking the following forms:
\begin{itemize}
    \item The identity function; This means our task-specific transform is just a low-rank linear transformation at each layer. To satisfy our parameter constraint we need $d_s=100$. We refer to this method as \textbf{Low-rank Layers}.
    \item Multi-head attention. To satisfy our parameter constraint we need $d_s=84$. We found that it was not necessary to use the $W^o$ matrix (see section~\ref{sec:param}) when adapting within BERT, and did not use it in any of our models.
    \item  Multi-head attention, with shared $V^E$ and $V^D$ across layers (not tasks). This parameter sharing allows a larger $d_s=204$. We refer to this method as \textbf{Projected Attention Layers (PALs)}.
    \item  Shared $V^E$ and $V^D$ across layers, but with $g(\cdot)$ a feedforward network with intermediate size 306 instead of attention (and again $d_s=204)$.
\end{itemize}
The motivation behind PALs is that we want to spend our parameter budget on transformations with an inductive bias useful for sequences. The `encoder' and `decoder' matrices operate on each sequence element separately, unlike attention, which transforms the input based on the entire sequence. Finally, the attention mechanism of PALs can potentially be inspected to see which tokens in a sequence the task-specific parts of the model focus on, although we did not concentrate on this aspect in this work.

\begin{table}[t]
\vspace*{-1.3ex}
\caption{How parameters are `spent' for some of our methods, where $T$ is the number of tasks, and there are 12 layers in the base network. The $2d_md_s$ terms come from `encoder' and `decoder' matrices. PALs (section~\ref{sec:within}) use $3d_s^2$ parameters per multi-head layer (see section~\ref{sec:param}) rather than $4d_s^2$ because they do not use the final linear transform $W^o$. Projected attention (section~\ref{sec:top}) worked best with six rather than twelve layers.}
\label{param-table}
\vskip 0.15in
\begin{center}
\begin{small}
\begin{sc}
\begin{tabular}{lcccccc|r}
\toprule
Method & Parameters \\
\midrule
PALs &  $T(2d_md_s$ + 12 $\times 3d_s^2$)\\
Low rank &  $T(12 \times2d_md_s$)\\
Proj.\ Attn.\ on top &  $T(2d_md_s + 6 \times4d_s^2$)\\
\bottomrule
\end{tabular}
\end{sc}
\end{small}
\end{center}
\end{table}

\section{Multi-task Training and Experiment Setup}
\subsection{Sampling Tasks}
\label{samp}
A simple way to train a model on several tasks is to select a batch of training examples from each task, cycling through them in a fixed order. We refer to this as `round-robin' sampling. However if the tasks have different numbers of training examples, round-robin sampling may not work well. By the time we have seen every example from a particular task we could have looped through another task's smaller dataset many times. This imbalance could lead to over-fitting on smaller tasks, and under-training on larger tasks. Potentially we could alleviate this issue by manually tuning regularisation hyper-parameters for each task.

Alternatively we can use methods where we see more examples from tasks with larger associated datasets. Concretely, we select a batch of examples from task $i$ with probability $p_i$ at each training step, and set $p_i$ proportional to $N_i$, the number of training examples for task $i$:
\begin{equation}
    p_i \propto N_i.
\end{equation}
This is the approach of the multi-task BiLSTM of \citet{glue} on the GLUE benchmark, and was used by \citet{hmtl}. It has the appealing property of selecting each example with the same probability as combining all the tasks and picking examples uniformly (though we train on batches from each task not single examples).

Since the ratio of the largest to the smallest task sizes $N_i$ we use is $\approx$158, we only rarely train on some tasks with the simple $\propto N_i$ method. Training on one task (or a particular subset of tasks) for many steps can lead to interference, where performance on the other tasks suffers.
A more general approach to sampling tasks sets $p_i$ as:
\begin{equation}
    p_i \propto N_i^\alpha.
\end{equation}
If we choose $\alpha < 1$ we reduce the disparity between the probabilities of choosing tasks. We consider $\alpha=0.5$ in our experiments, and call this method \textbf{`square root sampling'}.

Finally, we noticed that it was beneficial to train on tasks more equally towards the end of training, where we are most concerned about interference, and so we constructed the \textbf{`annealed sampling'} method where $\alpha$ changes with each epoch~$e$:
\begin{equation}
    \alpha = 1 - 0.8\frac{e-1}{E-1},
\end{equation}
where $E$ is the total number of epochs. Since we used multiple datasets we chose a somewhat arbitrary `epoch' of 2400 training steps.

It was particularly important to use the square root or annealed sampling methods when sharing a pooling layer (see section~\ref{sec:param}), and it makes intuitive sense that when the layer just before the output is shared, we need to guard against interference between tasks.
\subsection{Setup}
We based our experiments on the PyTorch implementation of BERT\,\footnote{https://github.com/huggingface/pytorch-pretrained-BERT} and open-source our code\footnote{https://github.com/AsaCooperStickland/Bert-n-Pals}.
No matter how we sampled tasks, we (unless stated otherwise) trained for 60,000 steps, with a minibatch size of 32, and a maximum sequence length of 128 tokens, choosing the best model from within that training time based on average development set score.
We
use Adam with learning rate of $2\!\times\!10^{-5}$, $\beta_1= 0.9$, $\beta_2= 0.999$, L2  weight  decay  of 0.01, learning
rate warmup over the first 10\% of steps (usually 6,000), and linear
decay of the learning rate after this, going down to zero at the end of training. We note warmup followed by linear decay is the `slanted triangular learning rate' of \citet{ulmft}, who find it is suited for fine-tuning a language model on single tasks. We performed most of our experiments using either the `proportional', `square root' or `annealed' sampling methods (see section~\ref{samp}). Round robin sampling gave consistently worse results.

We use twelve heads for the attention mechanism in PALs and other methods, except when using a smaller hidden size, where we decreased it proportionally. We did not find significant performance differences when changing the number of heads. We used the same BERT-base architecture as by \citet{bert}, twelve attention heads, $d_{ff}=3072$ and $d_m=768$ (see section~\ref{sec:param}).

We found it was crucial to use the pre-trained weights for BERT-base and not start from scratch. When training from scratch, with adaption parameters or not, we got significantly worse performance. For some tasks we did not get better results than random guessing after 90,000 steps. Although we note we used the same hyper-parameters as when training from the pre-trained weights, which might not be optimal for starting from scratch.
We experimented briefly with freezing the BERT-base parameters and fine-tuning only the PALs and alternatives, but concentrated on training all of the parameters, finding it took less parameters to approach matching fine-tuned BERT.

\subsection{Details of GLUE Tasks}
We test our methods for multi-task adaptation on eight of the nine tasks in the GLUE benchmark \citep{glue}\footnote{\citet{glue} provide a more detailed discussion of these tasks.}.

\textbf{Single-sentence tasks}:
Acceptability classification with
CoLA \citep{cola}; binary sentiment
classification with SST \citep{sst}.

\textbf{Sentence pair tasks}:
Semantic similarity with the MSR Paraphrase Corpus \citep[MRPC:][]{mrpc}, STS-Benchmark \citep[STS:][]{sts} and Quora
Question Pairs (QQP) dataset, and textual entailment with Multi-Genre NLI Corpus \citep[MNLI:][]{mnli}, a subset of the RTE challenge corpora \citep{rte}, and data from
SQuAD \citep[QNLI:][]{qnli}.

Like \citet{bert} we exclude the Winograd NLI task. When systems are trained on this task they have always performed worse than the 65.1
baseline accuracy of predicting the majority class. For our submissions we also simply predicted the majority class.
\section{Experiments and Discussion}
\begin{table*}[t]
\caption{GLUE Test results, scored by the GLUE evaluation server.  The number below each task denotes the
number of training examples. We show F1/accuracy scores for QQP and MRPC, and accuracy on the matched/mismatched test sets for MNLI\@. The `Av.'\ column is slightly different than the official GLUE score,  since
we  exclude WNLI\@. `Bert-base' results are from \citet{bert}. `Shared' refers to the model where all parameters are shared except the final projection to output space. The models we tested are a result of the `annealed sampling' method for multi-task training as it produced the best results on the dev set.}
\label{glue-table}
\vskip 0.15in
\begin{center}
\begin{small}
\begin{sc}
\scalebox{0.97}{\begin{tabular}{lccccccccc|r}
\toprule
Method & Params & MNLI-(m/mm) & QQP & QNLI & SST-2 & CoLA  & STS-B & MRPC & RTE & Av.\\
& & 392k & 363k& 108k & 67k & 8.5k & 5.7k & 3.5k & 2.5k \\
\midrule
BERT-base    & 8$\times$ & \underline{84.6}/83.4  & \underline{89.2}/71.2 & \underline{90.1} & \underline{93.5} & \underline{52.1} & \underline{85.8} & \underline{84.8}/\underline{88.9} & 66.4 & 79.6 \\
\midrule
Shared    & 1.00$\times$ & 84.0/83.4  & 88.9/70.8 & 89.3 & 93.4 & 51.2 & 83.6 & 81.3/86.7 & \underline{76.6} & 79.9 \\

Top Proj. Attn.     & 1.10$\times$ & 84.0/83.2  & 88.8/71.2 & 89.7& 93.2 & 47.1 & 85.3 & 83.1/87.5 & 75.5 & 79.6\\

PALs (204)    & 1.13$\times$ & 84.3/\underline{83.5}  & \underline{89.2}/\underline{71.5} & 90.0& 92.6 & 51.2 & \underline{85.8} & 84.6/88.7 & 76.0 & \textbf{80.4}\\

\bottomrule
\end{tabular}}
\end{sc}
\end{small}
\end{center}
\vskip -0.1in
\end{table*}

Table~\ref{glue-table} lists our results on GLUE for our best-performing PAL model (chosen by average development set performance), and some alternatives. Our main comparison is against fine-tuned BERT-base, which in the absence of transfer effects represents an upper bound on our performance, since it involves tuning all BERT-base parameters to perform well on each task individually, therefore requiring approximately 8$\times$ as many parameters as our methods. By construction, apart from our adaptation parameters we use the exact same architecture as BERT-base. We note that with the exception of our results for RTE, better performance can be obtained by fine-tuning the BERT-large model that has approximately 3$\times$ the parameters of BERT-base.

The use of multi-task training significantly improves results on the RTE task, achieving state-of-the-art performance. Similar improvements have been observed with multi-task LSTM-based systems \citep{glue} and by pre-training on MNLI before fine-tuning on RTE \citep{stilts}. Since RTE has the smallest number of training examples, and is similar to MNLI, it makes intuitive sense that it benefits from multi-task training. Sharing more parameters increased performance on RTE, and our fully-shared model has slightly better performance on RTE than PALs, however PALs are the only model that matches BERT-base on the larger tasks as well as performing well on RTE\@.

For the large sentence-pair tasks, MNLI, QQP and QNLI, performance is almost exactly the same as BERT-base with PALs.
For the two single
sentence tasks: the syntax-oriented CoLA task
and the SST sentiment task we see the largest drops in performance with PALs. This is in agreement with the results of \citet{stilts} who did not observe any transfer from various intermediate tasks, and, for CoLA, mirrors the results of \citet{elmfr} that language modeling alone is the best pre-training task for CoLA\@.

\begin{table*}[t]
\caption{GLUE performance, in terms of average score across each task's development set; this score is accuracy except for CoLA, where it is Matthews correlation, and STS-B, where it is Pearson correlation. We show the mean and standard error over three random seeds, unless standard error is $<0.005$. For the details of the sampling strategies see section~\ref{samp}. For the `within BERT' methods we show the smaller hidden state size in brackets, and write `no sharing' to refer to not sharing $V^E$ and $V^D$ across layers, `top' to mean adding in parallel to the six BERT layers just before the output, and `bottom' to mean adding in parallel to the six BERT layers just after the input. }
\label{top-table}
\vskip 0.15in
\begin{center}
\begin{small}
\begin{sc}
\begin{tabular}{lcccccr}
\toprule
Method & No.\ Params & New Layers  & Prop. Samp. & Sqrt.\ Samp. & Anneal Samp. \\
\midrule
Shared    & 1.00$\times$ & 0  & 79.17$\pm$0.03 & 80.56$\pm$0.04 & 80.7$\pm$0.3 \\

\midrule
Adding on top of BERT\\
\midrule
BERT Layer & 1.66$\times$ & 1 &80.6$\pm$0.2 & 81.6$\pm$0.3 & 81.5$\pm$0.2 \\
\midrule

Proj.\ Attn.       & 1.10$\times$& 6 & 80.3$\pm$0.1  & \underline{81.4$\pm$0.1} &    81.5$\pm$0.1      \\
Proj.\ FFN (1 layer)   & 1.10$\times$& 6 &  & 81.07 &   80.8$\pm$0.1      \\
\midrule
Adding within BERT\\
\midrule
PALs (204)       & 1.13$\times$& 12 & \underline{80.6$\pm$0.2} & 81.0$\pm$0.2 & \textbf{81.7$\pm$0.2}        \\
PALs no sharing (84)       & 1.13$\times$& 12 & &  & 81.3$\pm$0.1        \\
Low Rank (100)  & 1.13$\times$& 12 & &  &\textbf{81.9$\pm$0.2}        \\
PALs (276, top)   & 1.13$\times$& 6 &  &  & 81.61$\pm$0.06        \\
PALs (276, bottom)   & 1.13$\times$& 6 &  &  & 81.4$\pm$0.1        \\
\bottomrule
\end{tabular}
\end{sc}
\end{small}
\end{center}
\vskip -0.1in
\end{table*}

\subsection{PALs and Alternatives}
Table~\ref{top-table} lists our results on the GLUE benchmark development set for various ways of adding task-specific parameters and sampling strategies.

Our best results came with PALs, or low-rank layers, adapting every layer within BERT\@. The performance of PALs increased with a larger hidden state. Having separate `encoder' and `decoder' matrices (see section~\ref{sec:within}) across layers, or having separate pooling layers for each task, with the appropriate reduction in hidden state size to make up for the extra parameters, resulted in worse performance for PALs. However sharing `encoder' and `decoder' matrices between \textit{tasks} or both layers and tasks hurt results. A larger hidden state size seems important for Transformer models,
e.g.\ the performance of BERT-large vs.\ BERT-base \citep{bert} or the ablation study by \citet{NIPS2017_7181}.

We tested two adaption layers that did not use attention: Low-rank layers, and our method with shared `encoder' and `decoder' matrices but with a small feedforward network in-between them instead of attention. The latter model did not achieve good performance,
but low-rank layers and PALs have similar mean performance.

By inspecting the best-performing single models of each method we see a contrast: the strong results for low-rank layers are partly from better performance on CoLA\@. CoLA tends to see larger changes in score between models than other tasks since
it is scored by a different measure (Matthews correlation coefficient rather than accuracy). PALs performed better for the three largest tasks, MNLI, QQP and QNLI, and equivalently for other tasks.

These results suggest PALs has greater representational capacity; the only model that achieved comparable performance on the large tasks was adding an entire BERT-layer to the top, but this model had worse performance on the RTE task and uses many more parameters. The fact that spending parameters on linear transforms in the encoder, decoder or pooling matrices gives worse performance, and the worse performance of feedforward layers compared to multi-head attention, points towards the inductive bias provided by attention being important for good performance.

However at sufficiently parameter constrained regimes (for example 1.5 million parameters, which implies $d_s=10$ for low-rank transforms, and $d_s=60$ for PALs), PALs and low-rank layers performed similarly to the fully-shared model. Using the LHUC method (see section~\ref{adapt}), which requires even fewer parameters, also gave no improvement over the fully-shared baseline.

Ultimately, given the simplicity and competitive performance of low-rank layers, they remain an attractive option. There may be bigger differences for tasks like question answering which rely on the hidden states of every token in the input (as opposed to GLUE tasks which only use the final \texttt{[CLS]} hidden state to make predictions). We note that PALs and low-rank layers can easily be combined, say by using one type of adapter in the higher layers of the network and another in the lower ones.

When adding parameters to the top of BERT-base, it was important to use attention rather than feedforward transforms. Six additional layers worked best, outperforming using twelve or three layers. We also found it was crucial to use layer-norm and residual connections after each application of attention. Surprisingly, for these models using a separate pooling layer did not noticeably change results, and we report results with a shared pooling layer, which requires fewer parameters. These models saw worse performance on the RTE task, perhaps because transfer from other tasks is important, and splitting the model into multiple `heads' for each task dampens the benefits of shared knowledge.

\subsection{Where should we add Adaptation Modules?}

We draw some of the same conclusions as \citet{adapt} for `residual adapter modules'. As that work studied multi-task computer vision with residual networks (section~\ref{adapt}), we hope that these principles will apply broadly.

Adding task-specific functions \emph{within} networks works better than adding them to the \emph{top} (for a given number of parameters).  As found by \citet{adapt}, the best performing models had adaptations at \emph{every layer} of the base network, and adding adapter modules to the \emph{final half} of the base model worked better than adding to the half \emph{just after the input}. Unfortunately, adapting every layer of the base model represents the worst case for sharing operations between tasks. (We note again that this sharing is possible only when we want to perform many tasks on the same piece of text). But adapting the final half achieved slightly better performance than adding to the top of BERT-base. When adapting the final half we can still share the first six layers worth of operations, offering a useful compromise.

For within-network adaptations, \emph{parallel} connections worked better than \emph{serial} ones, also as found by \citet{adapt}. Our results with serial connections were much worse than simply not including any adapters. While the parallel configuration acts as a perturbation on the base network, the serial configuration more directly changes the hidden states being fed into the next layer. In these ways, the parallel configuration is less prone to the loss of the `knowledge' stored in the base network. We note that our serial configuration adds a newly initialised layer-norm, which may be the source of the performance drop.

\section{Further Discussion}

We found the details of how to schedule training examples from each task were important. With a lot of parameter sharing, sampling tasks proportional to dataset size impaired performance compared to our `annealing' method, where we slowly decrease the influence of dataset size on sampling probability. Annealing increased the variance of performance across random seeds as well as mean performance, meaning that we may need to pay the cost of several training runs to obtain the best single models from this method. We did not consider many variations of training method, and used no methods to reduce interference from training on separate tasks (to take one example, the `Gradient Episodic Memory' of \citealp{gem}). How these methods interact with choice of adaptation parameters is a direction for further research.

We introduced `Projected Attention Layers' as a transformation that can adapt the BERT sentence representation model for multi-task learning. PALs give a higher capacity for a given number of parameters compared to all the alternatives we considered, although simple low-rank transformations remain attractive due to their simplicity. If we adapt all the layers of BERT-base, we cannot share any operations across tasks. Ultimately the choice of which method to use depends on the constraints in place; if parameters are less constrained but you want to share as many operations as possible, adding an entire task-specific BERT layer on top of the model makes sense. If shared operations are not an issue, then adding PALs to every layer will perform well with few parameters. Finally, adapting only the final half of the base model offers a compromise between performance and sharing operations.

\section*{Acknowledgements}
We would like to thank Ivan Titov and Timothy Hospedales for useful discussion, and Elaine Farrow for help with a draft version of this paper. 
Asa Cooper Stickland was supported in part by the EPSRC Centre for Doctoral Training in Data Science, funded by the UK Engineering and Physical Sciences Research Council (grant EP/L016427/1) and the University of Edinburgh.
\bibliography{example_paper}

\begin{thebibliography}{38}
\providecommand{\natexlab}[1]{#1}
\providecommand{\url}[1]{\texttt{#1}}
\expandafter\ifx\csname urlstyle\endcsname\relax
  \providecommand{\doi}[1]{doi: #1}\else
  \providecommand{\doi}{doi: \begingroup \urlstyle{rm}\Url}\fi

\bibitem[Ba et~al.(2016)Ba, Kiros, and Hinton]{Ba2016LayerN}
Ba, J., Kiros, R., and Hinton, G.~E.
\newblock Layer normalization.
\newblock \emph{CoRR}, abs/1607.06450, 2016.

\bibitem[Bowman et~al.(2018)Bowman, Pavlick, Grave, Durme, Wang, Hula, Xia,
  Pappagari, McCoy, Patel, Kim, Tenney, Huang, Yu, Jin, and Chen]{elmfr}
Bowman, S.~R., Pavlick, E., Grave, E., Durme, B.~V., Wang, A., Hula, J., Xia,
  P., Pappagari, R., McCoy, R.~T., Patel, R., Kim, N., Tenney, I., Huang, Y.,
  Yu, K., Jin, S., and Chen, B.
\newblock Looking for {ELM}o's friends: Sentence-level pretraining beyond
  language modeling.
\newblock \emph{CoRR}, abs/1812.10860, 2018.

\bibitem[Caruana(1997)]{Caruana}
Caruana, R.
\newblock Multitask learning.
\newblock \emph{Mach. Learn.}, 28\penalty0 (1):\penalty0 41--75, July 1997.
\newblock ISSN 0885-6125.
\newblock \doi{10.1023/A:1007379606734}.

\bibitem[Cer et~al.(2017)Cer, Diab, Agirre, Lopez-Gazpio, and Specia]{sts}
Cer, D., Diab, M., Agirre, E., Lopez-Gazpio, I., and Specia, L.
\newblock Semeval-2017 task 1: Semantic textual similarity multilingual and
  crosslingual focused evaluation.
\newblock In \emph{Proceedings of the 11th International Workshop on Semantic
  Evaluation (SemEval-2017)}, pp.\  1--14. Association for Computational
  Linguistics, 2017.
\newblock \doi{10.18653/v1/S17-2001}.

\bibitem[Collobert et~al.(2011)Collobert, Weston, Bottou, Karlen, Kavukcuoglu,
  and Kuksa]{Collobert}
Collobert, R., Weston, J., Bottou, L., Karlen, M., Kavukcuoglu, K., and Kuksa,
  P.
\newblock Natural language processing (almost) from scratch.
\newblock \emph{J. Mach. Learn. Res.}, 12:\penalty0 2493--2537, November 2011.
\newblock ISSN 1532-4435.

\bibitem[Dagan et~al.(2006)Dagan, Glickman, and Magnini]{rte}
Dagan, I., Glickman, O., and Magnini, B.
\newblock The pascal recognising textual entailment challenge.
\newblock In \emph{Proceedings of the First International Conference on Machine
  Learning Challenges: Evaluating Predictive Uncertainty Visual Object
  Classification, and Recognizing Textual Entailment}, MLCW'05, pp.\  177--190,
  Berlin, Heidelberg, 2006. Springer-Verlag.
\newblock ISBN 3-540-33427-0, 978-3-540-33427-9.
\newblock \doi{10.1007/11736790_9}.

\bibitem[Dai \& Le(2015)Dai and Le]{dai}
Dai, A.~M. and Le, Q.~V.
\newblock Semi-supervised sequence learning.
\newblock In Cortes, C., Lawrence, N.~D., Lee, D.~D., Sugiyama, M., and
  Garnett, R. (eds.), \emph{Advances in Neural Information Processing Systems
  28}, pp.\  3079--3087. Curran Associates, Inc., 2015.

\bibitem[Dai et~al.(2019)Dai, Yang, Yang, Carbonell, Le, and Salakhutdinov]{xl}
Dai, Z., Yang, Z., Yang, Y., Carbonell, J.~G., Le, Q.~V., and Salakhutdinov, R.
\newblock Transformer-{XL}: Attentive language models beyond a fixed-length
  context.
\newblock \emph{CoRR}, abs/1901.02860, 2019.

\bibitem[Deng et~al.(2013)Deng, Hinton, and Kingsbury]{deng}
Deng, L., Hinton, G., and Kingsbury, B.
\newblock New types of deep neural network learning for speech recognition and
  related applications: an overview.
\newblock In \emph{2013 IEEE International Conference on Acoustics, Speech and
  Signal Processing}, pp.\  8599--8603, May 2013.
\newblock \doi{10.1109/ICASSP.2013.6639344}.

\bibitem[Devlin et~al.(2018)Devlin, Chang, Lee, and Toutanova]{bert}
Devlin, J., Chang, M., Lee, K., and Toutanova, K.
\newblock {BERT:} pre-training of deep bidirectional transformers for language
  understanding.
\newblock \emph{CoRR}, abs/1810.04805, 2018.

\bibitem[Dolan \& Brockett(2005)Dolan and Brockett]{mrpc}
Dolan, W.~B. and Brockett, C.
\newblock Automatically constructing a corpus of sentential paraphrases.
\newblock In \emph{Proceedings of the Third International Workshop on
  Paraphrasing (IWP2005)}, 2005.

\bibitem[Duong et~al.(2015)Duong, Cohn, Bird, and Cook]{duong}
Duong, L., Cohn, T., Bird, S., and Cook, P.
\newblock Low resource dependency parsing: Cross-lingual parameter sharing in a
  neural network parser.
\newblock In \emph{Proceedings of the 53rd Annual Meeting of the Association
  for Computational Linguistics and the 7th International Joint Conference on
  Natural Language Processing (Volume 2: Short Papers)}, pp.\  845--850.
  Association for Computational Linguistics, 2015.
\newblock \doi{10.3115/v1/P15-2139}.

\bibitem[Hashimoto et~al.(2017)Hashimoto, Xiong, Tsuruoka, and Socher]{hash}
Hashimoto, K., Xiong, C., Tsuruoka, Y., and Socher, R.
\newblock A joint many-task model: Growing a neural network for multiple nlp
  tasks.
\newblock In \emph{Proceedings of the 2017 Conference on Empirical Methods in
  Natural Language Processing}, pp.\  1923--1933. Association for Computational
  Linguistics, 2017.
\newblock \doi{10.18653/v1/D17-1206}.

\bibitem[He et~al.(2016)He, Zhang, Ren, and Sun]{resnet}
He, K., Zhang, X., Ren, S., and Sun, J.
\newblock Identity mappings in deep residual networks.
\newblock In \emph{ECCV}, 2016.

\bibitem[Hendrycks \& Gimpel(2016)Hendrycks and Gimpel]{gelu}
Hendrycks, D. and Gimpel, K.
\newblock Bridging nonlinearities and stochastic regularizers with gaussian
  error linear units.
\newblock \emph{CoRR}, abs/1606.08415, 2016.

\bibitem[{Houlsby} et~al.(2019){Houlsby}, {Giurgiu}, {Jastrzebski}, {Morrone},
  {de Laroussilhe}, {Gesmundo}, {Attariyan}, and {Gelly}]{bertadapt}
{Houlsby}, N., {Giurgiu}, A., {Jastrzebski}, S., {Morrone}, B., {de
  Laroussilhe}, Q., {Gesmundo}, A., {Attariyan}, M., and {Gelly}, S.
\newblock {Parameter-Efficient Transfer Learning for NLP}.
\newblock \emph{CoRR}, abs/1902.00751, 2019.

\bibitem[Howard \& Ruder(2018)Howard and Ruder]{ulmft}
Howard, J. and Ruder, S.
\newblock Universal language model fine-tuning for text classification.
\newblock In \emph{Proceedings of the 56th Annual Meeting of the Association
  for Computational Linguistics (Volume 1: Long Papers)}, pp.\  328--339.
  Association for Computational Linguistics, 2018.

\bibitem[Liu et~al.(2019)Liu, He, Chen, and Gao]{mtdnn}
Liu, X., He, P., Chen, W., and Gao, J.
\newblock Multi-task deep neural networks for natural language understanding.
\newblock \emph{CoRR}, abs/1901.11504, 2019.

\bibitem[Lopez-Paz \& Ranzato(2017)Lopez-Paz and Ranzato]{gem}
Lopez-Paz, D. and Ranzato, M.
\newblock Gradient episodic memory for continual learning.
\newblock In Guyon, I., Luxburg, U.~V., Bengio, S., Wallach, H., Fergus, R.,
  Vishwanathan, S., and Garnett, R. (eds.), \emph{Advances in Neural
  Information Processing Systems 30}, pp.\  6467--6476. Curran Associates,
  Inc., 2017.

\bibitem[McCann et~al.(2018)McCann, Keskar, Xiong, and Socher]{deca}
McCann, B., Keskar, N.~S., Xiong, C., and Socher, R.
\newblock The natural language decathlon: Multitask learning as question
  answering.
\newblock \emph{CoRR}, abs/1806.08730, 2018.

\bibitem[Phang et~al.(2018)Phang, F{\'{e}}vry, and Bowman]{stilts}
Phang, J., F{\'{e}}vry, T., and Bowman, S.~R.
\newblock Sentence encoders on {STILTS}s: Supplementary training on
  intermediate labeled-data tasks.
\newblock \emph{CoRR}, abs/1811.01088, 2018.

\bibitem[Radford(2018)]{gpt}
Radford, A.
\newblock Improving language understanding by generative pre-training.
\newblock 2018.

\bibitem[Rajpurkar et~al.(2016)Rajpurkar, Zhang, Lopyrev, and Liang]{qnli}
Rajpurkar, P., Zhang, J., Lopyrev, K., and Liang, P.
\newblock Squad: 100,000+ questions for machine comprehension of text.
\newblock In \emph{Proceedings of the 2016 Conference on Empirical Methods in
  Natural Language Processing}, pp.\  2383--2392. Association for Computational
  Linguistics, 2016.
\newblock \doi{10.18653/v1/D16-1264}.

\bibitem[Rebuffi et~al.(2018)Rebuffi, Bilen, and Vedaldi]{adapt}
Rebuffi, S.-A., Bilen, H., and Vedaldi, A.
\newblock Efficient parametrization of multi-domain deep neural networks.
\newblock In \emph{IEEE Conference on Computer Vision and Pattern Recognition}.
  IEEE, 2 2018.

\bibitem[Ruder(2017)]{ruder}
Ruder, S.
\newblock An overview of multi-task learning in deep neural networks.
\newblock \emph{CoRR}, abs/1706.05098, 2017.

\bibitem[Sanh et~al.(2018)Sanh, Wolf, and Ruder]{hmtl}
Sanh, V., Wolf, T., and Ruder, S.
\newblock A hierarchical multi-task approach for learning embeddings from
  semantic tasks.
\newblock \emph{CoRR}, abs/1811.06031, 2018.

\bibitem[Socher et~al.(2013)Socher, Perelygin, Wu, Chuang, Manning, Ng, and
  Potts]{sst}
Socher, R., Perelygin, A., Wu, J., Chuang, J., Manning, C.~D., Ng, A., and
  Potts, C.
\newblock Recursive deep models for semantic compositionality over a sentiment
  treebank.
\newblock In \emph{Proceedings of the 2013 Conference on Empirical Methods in
  Natural Language Processing}, pp.\  1631--1642. Association for Computational
  Linguistics, 2013.

\bibitem[Subramanian et~al.(2018)Subramanian, Trischler, Bengio, and Pal]{sub}
Subramanian, S., Trischler, A., Bengio, Y., and Pal, C.~J.
\newblock Learning general purpose distributed sentence representations via
  large scale multi-task learning.
\newblock In \emph{International Conference on Learning Representations}, 2018.

\bibitem[Swietojanski \& Renals(2014)Swietojanski and Renals]{7078569}
Swietojanski, P. and Renals, S.
\newblock Learning hidden unit contributions for unsupervised speaker
  adaptation of neural network acoustic models.
\newblock In \emph{2014 IEEE Spoken Language Technology Workshop (SLT)}, pp.\
  171--176, Dec 2014.
\newblock \doi{10.1109/SLT.2014.7078569}.

\bibitem[Teh et~al.(2017)Teh, Bapst, Czarnecki, Quan, Kirkpatrick, Hadsell,
  Heess, and Pascanu]{rein}
Teh, Y., Bapst, V., Czarnecki, W.~M., Quan, J., Kirkpatrick, J., Hadsell, R.,
  Heess, N., and Pascanu, R.
\newblock Distral: Robust multitask reinforcement learning.
\newblock In Guyon, I., Luxburg, U.~V., Bengio, S., Wallach, H., Fergus, R.,
  Vishwanathan, S., and Garnett, R. (eds.), \emph{Advances in Neural
  Information Processing Systems 30}, pp.\  4496--4506. Curran Associates,
  Inc., 2017.

\bibitem[Vaswani et~al.(2017)Vaswani, Shazeer, Parmar, Uszkoreit, Jones, Gomez,
  Kaiser, and Polosukhin]{NIPS2017_7181}
Vaswani, A., Shazeer, N., Parmar, N., Uszkoreit, J., Jones, L., Gomez, A.~N.,
  Kaiser, L.~u., and Polosukhin, I.
\newblock Attention is all you need.
\newblock In Guyon, I., Luxburg, U.~V., Bengio, S., Wallach, H., Fergus, R.,
  Vishwanathan, S., and Garnett, R. (eds.), \emph{Advances in Neural
  Information Processing Systems 30}, pp.\  5998--6008. Curran Associates,
  Inc., 2017.

\bibitem[Wang et~al.(2018{\natexlab{a}})Wang, Singh, Michael, Hill, Levy, and
  Bowman]{glue}
Wang, A., Singh, A., Michael, J., Hill, F., Levy, O., and Bowman, S.
\newblock {GLUE}: A multi-task benchmark and analysis platform for natural
  language understanding.
\newblock In \emph{Proceedings of the 2018 EMNLP Workshop BlackboxNLP:
  Analyzing and Interpreting Neural Networks for NLP}, pp.\  353--355.
  Association for Computational Linguistics, 2018{\natexlab{a}}.

\bibitem[Wang et~al.(2018{\natexlab{b}})Wang, Girshick, Gupta, and
  He]{NonLocal2018}
Wang, X., Girshick, R., Gupta, A., and He, K.
\newblock Non-local neural networks.
\newblock \emph{CVPR}, 2018{\natexlab{b}}.

\bibitem[Warstadt et~al.(2018)Warstadt, Singh, and Bowman]{cola}
Warstadt, A., Singh, A., and Bowman, S.~R.
\newblock Neural network acceptability judgments.
\newblock \emph{CoRR}, abs/1805.12471, 2018.

\bibitem[Williams et~al.(2018)Williams, Nangia, and Bowman]{mnli}
Williams, A., Nangia, N., and Bowman, S.
\newblock A broad-coverage challenge corpus for sentence understanding through
  inference.
\newblock In \emph{Proceedings of the 2018 Conference of the North American
  Chapter of the Association for Computational Linguistics: Human Language
  Technologies, Volume 1 (Long Papers)}, pp.\  1112--1122. Association for
  Computational Linguistics, 2018.
\newblock \doi{10.18653/v1/N18-1101}.

\bibitem[Yang \& Hospedales(2017)Yang and Hospedales]{yang}
Yang, Y. and Hospedales, T.~M.
\newblock Trace norm regularised deep multi-task learning.
\newblock In \emph{ICLR Workshop}, 2017.

\bibitem[Zellers et~al.(2018)Zellers, Bisk, Schwartz, and
  Choi]{zellers2018swagaf}
Zellers, R., Bisk, Y., Schwartz, R., and Choi, Y.
\newblock Swag: A large-scale adversarial dataset for grounded commonsense
  inference.
\newblock In \emph{Proceedings of the 2018 Conference on Empirical Methods in
  Natural Language Processing (EMNLP)}, 2018.

\bibitem[Zhang et~al.(2018)Zhang, Goodfellow, Metaxas, and Odena]{sagan}
Zhang, H., Goodfellow, I.~J., Metaxas, D.~N., and Odena, A.
\newblock Self-attention generative adversarial networks.
\newblock \emph{CoRR}, abs/1805.08318, 2018.

\end{thebibliography}
\bibliographystyle{icml2019}

\clearpage
\appendix

\section{Performance on Tasks Over Time}
\label{submission}

\begin{figure*}[t]%
    \centering
    \subfloat[PALs]{{\includegraphics[width=8.1cm]{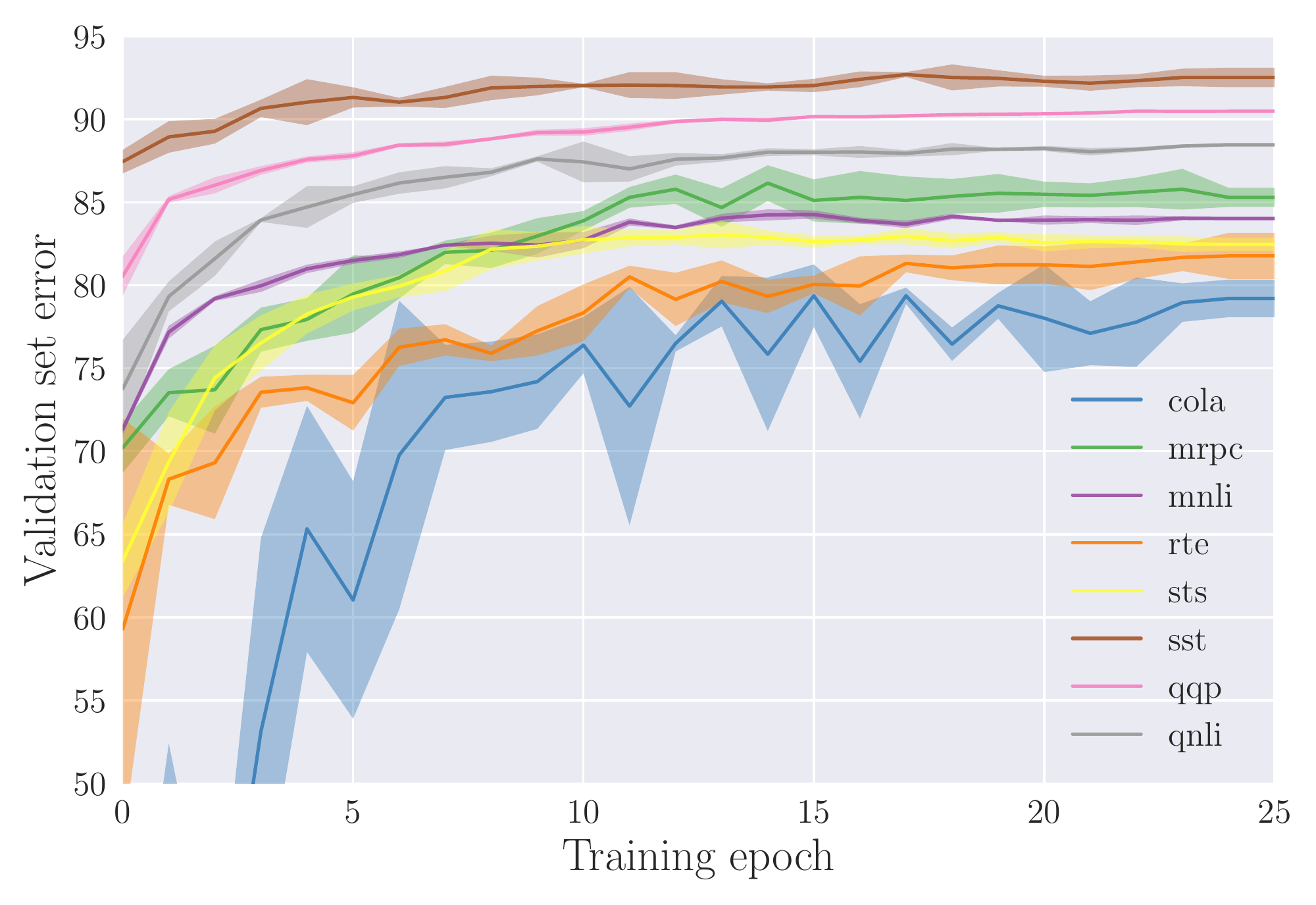} }}%
    \qquad
    \subfloat[Low rank adapters]{{\includegraphics[width=8.1cm]{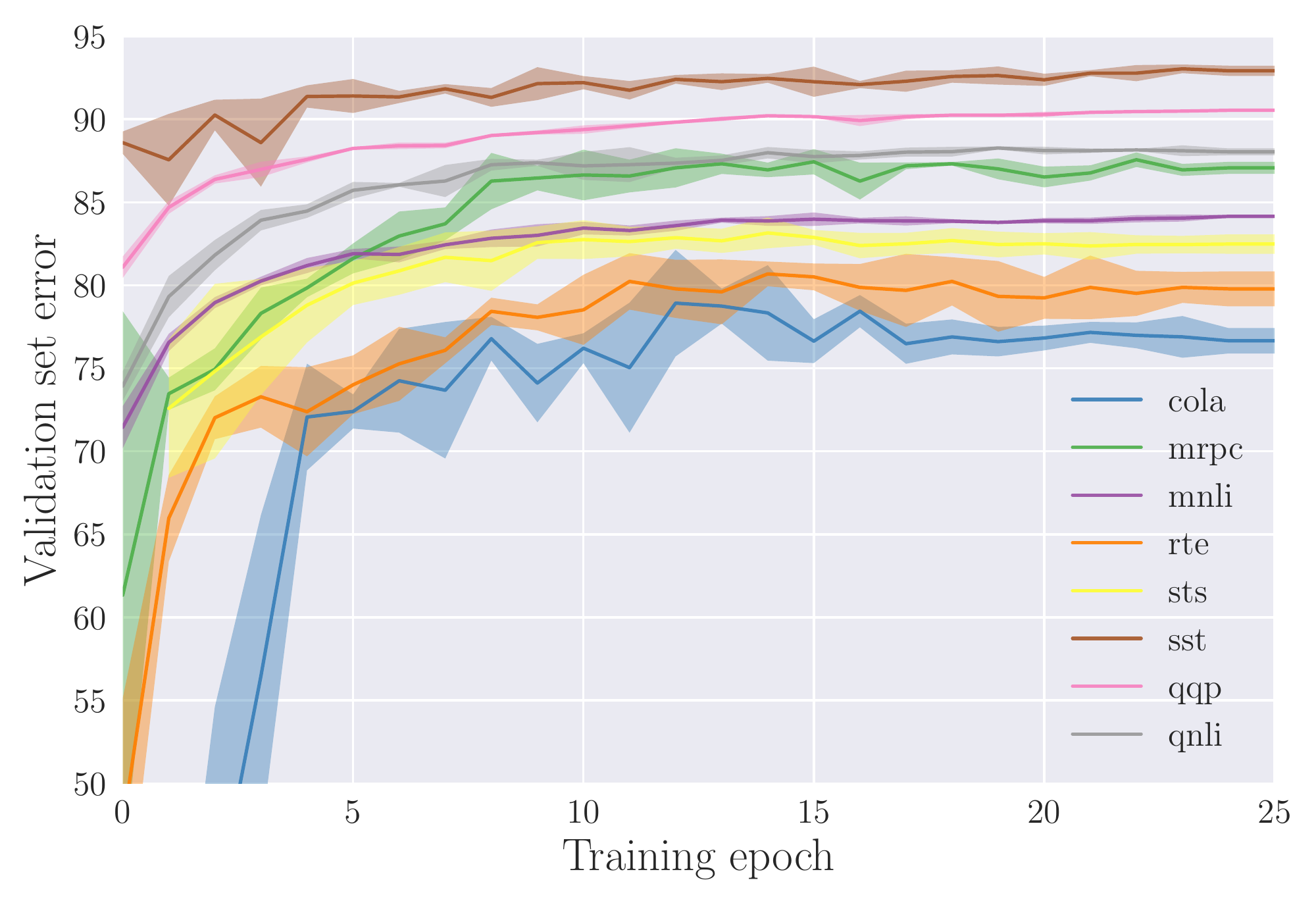} }}%
    \caption{Average performance over four random seeds for two adapter modules, with the shaded region indicating standard deviation. CoLA performance has been shifted up by 30\% for visibility.}%
    \label{fig:example}%
\end{figure*}

Figure~\ref{fig:example} shows performance on the GLUE tasks over time for PALs and low-rank adapter modules. The low-resource tasks have a much larger variation in performance than the high resource ones, which are fairly stable.
CoLA performance in particular varies a lot early on in training. Performance on CoLA and RTE goes down towards the end of training with low-rank adapters, and not with PALs, and the opposite trend for MRPC\@. These downward trends might be rectified with a better training schedule or regularisation scheme.

\section{Squad and SWAG Performance}
We conducted limited experiments on two additional tasks. The Stanford Question Answering Dataset (SQuAD) is a collection of 100k crowdsourced question/answer pairs \cite{qnli}, where the task is to predict the location of the answer in a paragraph from Wikipedia. We follow the approach of \citet{bert} by associating each token in the input sequence with a probability of being the start, and end, of the answer span. The Situations With Adversarial Generations (SWAG) dataset contains 113k sentence-pair completion examples intended to evaluate grounded commonsense inference \cite{zellers2018swagaf}. Given a sentence from a video captioning dataset, the task is to decide among four choices the most plausible continuation, with each sentence-completion pair assigned a score, and a softmax applied over the four choices to form a probability distribution.

We tested multi-task learning with the SQuAD and SWAG datasets. We follow all the same experimental settings as before, but we use round robin sampling because of the comparable size of the datasets, and train for 24,000 steps, not 60,000, with an increased maximum sequence length, 256. Results, see table~\ref{top-table}, show a slight improvement when using the PAL adapters compared to a fully shared baseline and low-rank adapters. However all approaches performed similarly, with there perhaps less need for the flexibility provided by adapters when only training on two tasks.

\begin{table*}[t]
\caption{Performance on SQuAD and SWAG, in terms of average score across each task's development set; this score is exact match and f1 score for SQuAD, and accuracy for SWAG.}
\label{top-table}
\vskip 0.15in
\begin{center}
\begin{small}
\begin{sc}
\begin{tabular}{lcccr}
\toprule
Method & No.\ Params & New Layers  & Round Robin \\
\midrule
Shared    & 1.00$\times$ & 0  & 82.75$\pm$0.09 \\
        \\
\midrule
Adding within BERT\\
\midrule
PALs (204)       & 1.13$\times$& 12 &  82.774$\pm$0.006     \\

Low Rank (100)  & 1.13$\times$& 12 &    82.74$\pm$0.06  \\

\bottomrule
\end{tabular}
\end{sc}
\end{small}
\end{center}
\vskip -0.1in
\end{table*}

\end{document}